\documentclass[journal]{IEEEtran}
\usepackage{graphicx, subfigure}
\usepackage{enumerate}
\usepackage{amsmath,amssymb} 
\usepackage{color}
\usepackage{changes}
\usepackage{tabularx}
\usepackage{multirow}
\usepackage[linguistics, edges]{forest}
\usetikzlibrary{arrows.meta,shadows.blur}

\usepackage[space]{cite}

\ifCLASSINFOpdf
\else
\fi
\usepackage{amsmath}
\usepackage{url}

\begin{document}
%
\title{ Anomaly Detection in Road Traffic Using Visual Surveillance: A Survey}
%
%
%

\author{Kelathodi~Kumaran~Santhosh,~\IEEEmembership{Student Member,~IEEE,}
        Debi~Prosad~Dogra,~\IEEEmembership{Member,~IEEE,} and
        Partha~Pratim~Roy
\thanks{K.~K.~Santhosh and D.~P.~Dogra are with School of Electrical Sciences, Indian Institute of Technology Bhubaneswar, Odisha,
India e-mail: (sk47@iitbbs.ac.in, dpdogra@iitbbs.ac.in).}
\thanks{P.~P.~Roy is with the Department of Computer Science and Engineering, Indian Institute of Technology, Roorkee, India. e-mail:(proy.fcs@iitr.ac.in).}
}

%
%

\markboth{preprint}%
{Shell \MakeLowercase{\textit{et al.}}: Bare Demo of IEEEtran.cls for IEEE Journals}
%



\maketitle

\begin{abstract}
 Computer vision has evolved in the last decade  as a key technology for numerous applications replacing human supervision. In this paper, we present a survey on relevant visual surveillance related researches for anomaly detection in public places, focusing primarily on roads. Firstly, we revisit the surveys done in the last 10 years in this field. Since the underlying building block of a typical anomaly detection is learning, we emphasize more on learning methods applied on video scenes. We then summarize the important contributions made during last six years on anomaly detection primarily focusing on features, underlying techniques, applied scenarios and types of anomalies using single static camera. Finally, we discuss the challenges in the computer vision related anomaly detection techniques and some of the important future  possibilities.
\end{abstract}


%
\IEEEpeerreviewmaketitle

\section{Introduction}
\IEEEPARstart{W}{ith} the widespread use of surveillance cameras in public places, computer vision-based scene understanding has gained a lot of popularity amongst the CV research community. Visual data contains rich information compared to other information sources such as GPS, mobile location, radar signals, etc. Thus, it can play a vital role in detecting/predicting congestions, accidents and other anomalies apart from collecting statistical information about the status of road traffic.
\begin{figure}[!ht]
\centering
\includegraphics[width=0.48\textwidth]{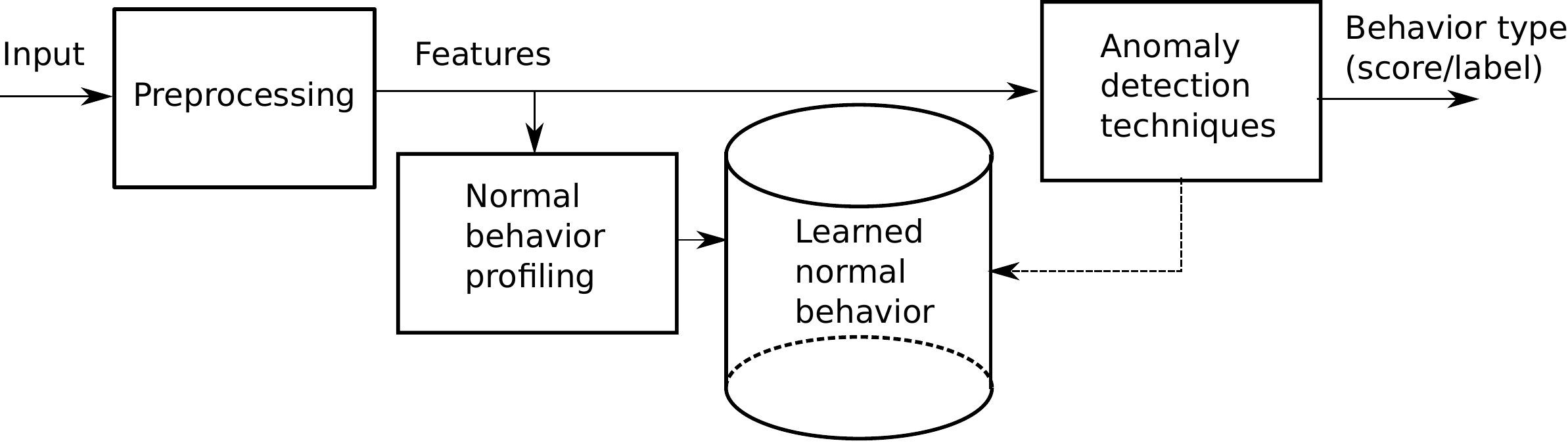}
\caption{Overview of a typical anomaly detection scheme. Preprocessing block extracts features/data in the form of descriptors. The normal behavior is represented in abstract form in terms of rules, models, or data repository. Specific anomaly detection techniques are used for detecting anomalies using anomaly scoring or labeling mechanism.}
\label{Fig:Scheme}
\end{figure}
\begin{figure*}[!h]
\centering
\includegraphics[width=1.0\textwidth]{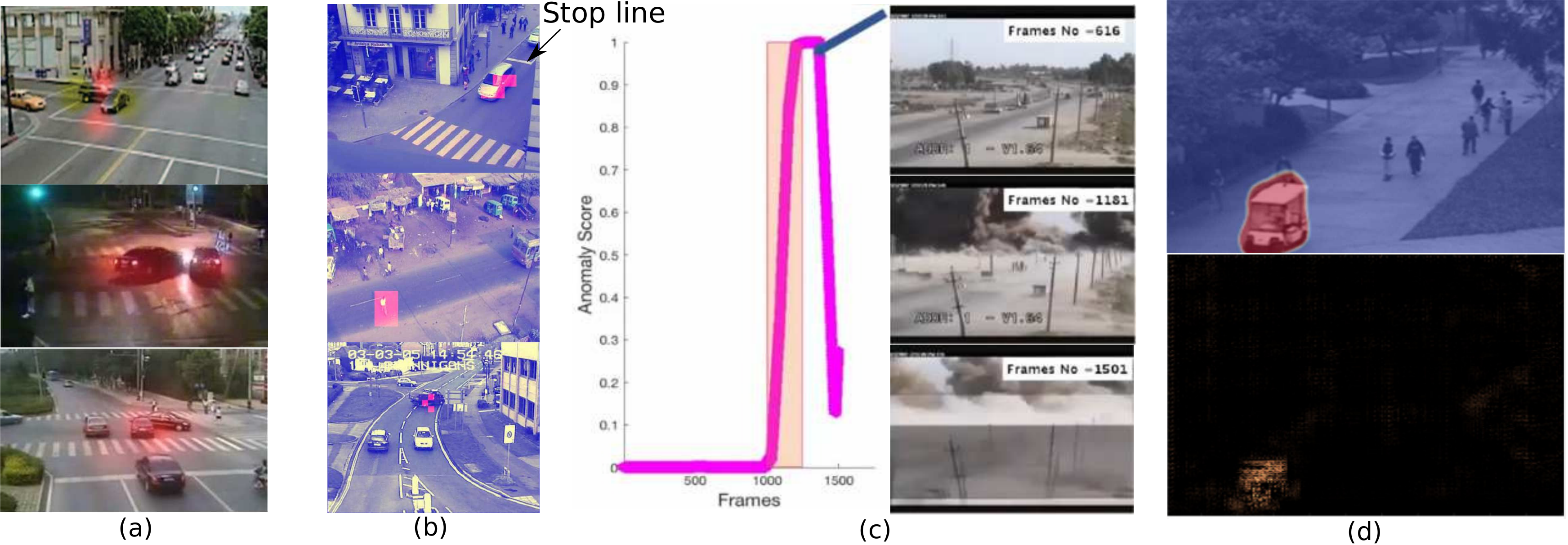}
\caption{Visual snapshots of some of the state-of-art anomaly detection techniques to present an overview about the survey. (a)  Accident detection using Motion Interaction Field (MIF)~\cite{2014_Yun_motion_MIF_Heuristic}. (b) Anomaly detection using topic-based models~\cite{2015_pathak_Feature_HOG_HOF_pLSA_anomaly_projectionmodel_histogrambased}.  Top row shows a vehicle that crossed the stop line, middle row represents a jaywalking scenario and the bottom row represents a vehicle taking unusual turn. (c) Real world anomaly detection using multiple instance learning (MIL)~\cite{sultani2018real_supervised}. The anomaly detection is measured using an anomaly score in explosion scene. (d) Presence of a vehicle on a walkway detected using spatio-temporal adversarial networks (STAN)~\cite{2018_lee_stan_Supervised_input_realandfakesequence_ModelGAN_Anomaly_deviation}. The top row represents the anomaly visualization from the generator and the bottom row represents the anomaly visualization from the discriminator. }
\label{Fig:Anomaly_demo}
\end{figure*}
Several computer vision-based studies have been conducted focusing on  data acquisition~\cite{2015_B_Tian}, feature extraction~\cite{2014_kantorov_Feature, 2012_AA_Sodemann}, scene learning~\cite{2008_BT_Morris, 2014_Bastani_Learning, 2013_Hu_Learning, 2015_YS_Chong}, activity learning~\cite{2013_Vishwakarma}, behavioral understanding~\cite{2013_Sivaraman_looking, 2015_Bastani_BehaviorAnalysis}, etc. These studies primarily discuss on aspects such as scene analysis, video processing techniques, anomaly detection methods, vehicle detection and tracking, multi camera-based techniques and challenges, activity recognition, traffic monitoring, human behavior analysis, emergency management, event detection, etc.  	

Anomaly detection is a sub-domain of behavior understanding~\cite{2015_B_Tian} from surveillance scenes. Anomalies are typically aberrations of scene entities (vehicles, human or the environment) from the normal behavior. With the availability of video feeds from public places, there has been a surge in the research outputs on video analysis and anomaly detection~\cite{2013_Sivaraman_looking, 2012_AA_Sodemann, 2017_MS_Shirazi, 2018_Mabrouk_abnormal}. Typically anomaly detection methods learn the normal behavior via training. Anything deviating significantly from the normal behavior can be termed as anomalous. Vehicle presence on walkways, a sudden dispersal of people within a gathering, a person falling suddenly while walking, jaywalking, signal bypassing at a traffic junction, or U-turn of vehicles during red signals are a few examples of anomalies. Anomaly detection frameworks typically use unsupervised, semi-supervised or unsupervised learning. In this survey, we mainly  explore anomaly detection techniques used in road traffic scenarios focusing on \textit{entities} such as vehicles, pedestrian, environment  and their interactions. 

We have noted that scope of the study should cover the nature of input data and their representations, feasibility of supervised learning, types of anomalies, suitability of the techniques in application contexts, anomaly detection outputs and evaluation criteria. We present this survey from the above perspectives. A typical anomaly detection framework is presented in Fig.~\ref{Fig:Scheme}. Usually, anomaly detection systems work by learning the normal data patterns to build a normal profile. Once the normal patterns are learned, anomalies can be detected with the help of established approaches~\cite{omar2013machine, li2016anomaly}. Output of the system can be a score typically in the form of a metric or a label that notifies whether the data is anomalous or not. 

 Some examples of anomaly detection results are shown in Fig.~\ref{Fig:Anomaly_demo}.

\subsection{Recent Surveys}
During last 10 years or so, a few interesting surveys have been published in this field of research.
Authors of~\cite{2008_BT_Morris} have explored object detection, tracking, scene modeling and activity analysis using video trajectories.  The study presented in~\cite{2011_B_Tian} covers vehicle  detection, tracking, behavior understanding and incident detection from the purview of intelligent transportation systems (ITS). Authors of \cite{2011_N_Buch} have conducted an in-depth study of traffic analysis frameworks under different taxonomies  with pointers at integrating information from multiple sensors. The review presented in \cite{2012_AA_Sodemann} is possibly the first work covering anomaly detection techniques. It covers sensors, entities, feature extraction methods, learning methods and scene modeling to detect anomalies. In~\cite{2013_Sivaraman_looking}, an object oriented approach  from  the perspective of vehicle mounted sensors for object detection, tracking and behavior analysis detailing the progress of the last decade of works, has been presented. Multi-camera study presented in~\cite{2013_X_Wang} covers the researches related to surveillance in  multi-camera setups. Authors of~\cite{2013_Suriani} discuss events, which are considered as a subset of anomalous events, requiring immediate attention, occuring unintentionally, abruptly and unexpectedly. The research presented in \cite{2013_Lose} discusses safety, security and law enforcement related applications from the computer vision perspective.  The review presented in \cite{2013_Vishwakarma} discusses the elements of human activity and behavioral understanding frameworks. Authors of~\cite{2013_PVK_Borges} present the researches on human behavioral understanding through actions and interactions of human entities.  Intelligent video systems covering analytics aspect  has been studied in \cite{2013_H_Liu}. Surveillance systems with specific application areas have been presented in \cite{2014_M_Zablocki}. Authors of \cite{2015_B_Tian} systematically divide road traffic analysis into four layers, namely image acquisition, dynamic and static attribute extraction, behavioral understanding and ITS services. Datasets used for anomaly detections have been covered in \cite{2016_N_Patil}. Traffic monitoring using different types of sensors has been discussed in \cite{2016_SRE_Datondji}. Algorithms used for spatio-temporal point detections and their applications in vision domain have been covered in \cite{2017_Y_Li}. Traffic entities have been studied from the perspective of safety in \cite{2017_MS_Shirazi}. Authors of \cite{2018_SA_Ahmed} explore studies on video trajectory-based analysis and applications. Authors of \cite{2018_L_Lopez-Fuentes} discuss various ways of handling  emergency situations by assessing the risks, preparedness, response, recovery and mitigation using the extracted information from the visual features with the help of various learning mechanisms. In~\cite{2018_Mabrouk_abnormal}, authors have presented anomalous human behavior recognition work with focus on behavior representation and modeling, feature extraction techniques, classification and behavior modeling frameworks, performance evaluation techniques, and datasets with examples of video surveillance systems. Table~\ref{Tab:Survey} summarizes the major computer vision-based studies done during last 10 years. In our survey, we particularly focus on the studies on anomaly detection that are relevant on road traffic scenarios.   

 Anomalies are contextual in nature. The assumptions used in anomaly detections cannot be applied universally across different traffic scenarios. We analyze the capabilities of anomaly detection methods used in road traffic surveillance from the perspective of data. In the process, we categorize the methods according to scene representation, employed features, used models and approaches.

\begin{table*}[!h]
\scriptsize
\begin{center}
\caption {Surveys on computer vision-based  methods in surveillance} 
\label{Tab:Survey} 
    \begin{tabular}{@{}p{0.06\textwidth}  p{0.12\textwidth}  p{0.78\textwidth}@{}}
\hline
    \textbf{Ref.} & \textbf{Focus} & \textbf{Explored research areas} \\ 
\hline
Morris (2008) \cite{2008_BT_Morris} 
		& Video trajectory-based scene analysis  
				&Scene modeling: Tracking, interest point study, activity path learning; Applications: People movement, traffic, parking lot, and entity interaction;  Path learning: Preprocessing (normalization and dimensionality reduction), clustering approaches and used distance measures, path modeling, relevance of path feedback in low level systems;  
Activity analysis: Virtual fencing, speed profiling, path classification, abnormality detection, online activity analysis, object interaction characterization.\\ 
\hline
Tian (2011) \cite{2011_B_Tian} 
		& Video processing techniques applied for traffic monitoring 
				&Traffic parameters collection; Traffic incident detection; Vehicle detection scenarios:  Background modeling and non-background modeling  approaches, shadow detection and removal; Vehicle tracking, model-based classification, region, deformable template and feature study, tracking algorithms; Traffic incident detection and behavior understanding.\\ 
\hline
Buch (2011) \cite{2011_N_Buch}) 
				& Video analytics system for urban traffic
						& Applications: Vehicle counting, automatic number plate recognition, incident detection; Analytics system components; Foreground segmentation techniques: Frame differencing, background subtraction (averaging, single Gaussian, mode estimation, Kalman filter, wavelets), GMM, graph cuts, shadow removal, object-based segmentation; Top-down vehicle classification: Features (region based, contour based), machine learning techniques; Bottom-up approaches: Interest point descriptors, object classification;  Tracking: Kalman filter, PF, S-T MRF, graph correspondence, event cones; Traffic analytic system: Urban (camera domain, three dimensional modeling), highways (detection and classification). 
							\\ 
\hline
Sodemann (2012) \cite{2012_AA_Sodemann} 
				& Anomaly detection 
							& Study on sensors: Visible-spectrum camera (low-level feature extraction and object level feature extraction), audio and infrared sensors; Learning methods: Unsupervised, supervised and apriori modeling; Classification algorithms: Dynamic bayesian networks, bayesian topic models, artificial neural networks, clustering, decision trees, fuzzy reasoning. \\ 
\hline
Sivaraman (2013)~\cite{2013_Sivaraman_looking}      
				& Vision-based vehicle detection, tracking and behavior analysis
							& Sensors: radar, lidar, camera;  Vehicle detection: Monocular vision (camera placement, appearance features and classification, motion based approaches, vehicle pose). Stereo vision (matching, motion-based approaches); Vehicle tracking: Monocular and stereo tracking, vision cue fusion, real-time challenges and system architecture, fusion with other modalities; Behavior analysis: context,  vehicle maneuvers, trajectories, behavioral classification; Future direction of vehicle detection, tracking, their on-road behavior and public benchmarks.\\
\hline
Wang (2013) \cite{2013_X_Wang} 
				& Multi-camera based surveillance
							&Multi-camera calibration; Topology computation; Multi-camera object tracking: Calibration, appearance cues, correspondence-based methods; Object re-identification: Feature studies, learning methods; Multi-camera activity analysis: Correspondence free methods, activity models, human action recognition; Cooperative video surveillance using active and static cameras; Background modeling and object tracking with active cameras.	\\      
\hline    
Suriani (2013) \cite{2013_Suriani} 
				& Abrupt event detection 
							& Human centered, vehicle centered and small area centered studies; Methods of detection: Single person, multiple person, vehicles, multi-view camera based.	 \\    
\hline  
Loce (2013) \cite{2013_Lose} 
				& Traffic management
							&Vehicle mounted camera-based safety applications: Lane departure warning and lane change assistance, pedestrian detection, driver monitoring, adaptive warning systems; Efficiency studies: Traffic flow management, incident management, video based tolling; Security management: Alert and warning systems, traffic surveillance, recognizing and tracking vehicles of interest; Law enforcement: Studies on speed enforcement, violation detection at road intersections,  vehicle mounted mobile camera based vehicle identification.\\  
\hline   
Vishwakarma (2013) \cite{2013_Vishwakarma} 
				& Human activity recognition and behavior analysis
							& Application areas: Behavioral biometrics, content-based video analysis, security and surveillance, interactive applications, animation and synthesis; Object detection methods: Motion segmentation methods (background subtraction based, statistical, temporal differencing and optical flow-based) and object classification; Object tracking methods (region, contour, feature, model, hybrid and optical flow-based); Action recognition techniques: Hierarchical (statistical, syntactic and description based) and non-hierarchical approaches; Human behavior understanding: Supervised, semi-supervised and unsupervised models; Dataset description: Controlled and realistic environments and its realistic impact on video-based surveillance market. 
										 \\  
\hline  
Borges (2013) \cite{2013_PVK_Borges} 
				& Human behavior analysis 
							& Human detection methods: Appearance, motion and hybrid approaches; Action recognition approaches: Low-level and spatio-temporal interest points, mid and high-level, silhouettes features; Interaction recognition: One-to-one, group interactions, models; Datasets.
										 \\  
\hline    
Liu (2013) \cite{2013_H_Liu} 
				& Intelligent video systems and analytics
							& Video systems:  Architecture (distributed/centralized), quality diagnosis, system adaptability (configuration, calibration, capability and scalability) analysis, data management and transmission methods; Analytics:  Object attributes, motion pattern recognition, event and behavior analysis; Analytic methods: Intelligence and cooperative aspects, multi-camera view selections, statistical and networked analysis, learning and classification, 3-D sensing; Applications areas:  Management, traffic control, transportation, intelligent vehicles, health-care, life sciences, security and military. \\  
\hline   
Zablocki (2013) \cite{2014_M_Zablocki} 
				& Characteristics of intelligent video surveillance systems 
							& System classification: Object detection, tracking and movement analysis technologies; Anomaly detection, identification and warning/alarming systems; Vehicle detection, traffic and parking lot analysis systems; Object counting systems; Integrated camera view    handling systems;  Privacy preserving systems; Cloud-based systems.	\\  
\hline   
Tian (2015) \cite{2015_B_Tian} 
				& Vehicle surveillance
							& Dynamic and static attribute extraction:  Appearance and motion-based detection, tracking, recognition (license plate, type, color and logo), networked tracking of vehicles; Behavior understanding:  Single camera study, trajectory (clustering, modeling and retrieval) and networked multi-camera-based, interesting region discovery; Image acquisition: Traffic scene characteristics, imaging technologies; ITS service study: Illegal activity and anomaly detection, security monitoring, electronic toll collection, traffic flow analysis, transportation planning and road construction, environment impact assessment.	\\ 
\hline
Patil (2016) \cite{2016_N_Patil} 
				& Video datasets for anomaly detection
							& Dataset classification: Traffic, subway, panic driven, pedestrian, abnormal activity, campus, train, sea, crowd.	\\     
\hline
Datondji (2016) \cite{2016_SRE_Datondji} 
				& Traffic monitoring at intersections 
							&  Camera based classification: Mono vision, omni vision and stereo vision; Vehicle sensing: Methodologies and datasets; Challenges: Initialization and preprocessing, vehicle detection and tracking; Vehicle detection methods:  Candidate localization, verification; Vehicle tracking: Representation and tracking approaches: Region, contour, feature and model-based; Vehicle tracking algorithms: Matching, Bayesian; Challenges for intersection; Monitoring systems:  Monocular vision and omni-directional vision-based, in-vehicle monitoring; Vehicle tracking: Roadside monitoring systems, in-vehicle monitoring systems; Vehicle behavior analysis. \\     
\hline  
Li (2017) \cite{2017_Y_Li} 
				& Spatio-temporal interest point (STIP) detection algorithms  
							&  STIPs algorithms; Detection challenges;  Applications: Human activity detection, anomaly detection, video summarization and content based video retrieval. \\     
\hline  
Shirazi (2017) \cite{2017_MS_Shirazi} 
				& Intersections analysis from safety perspective  
							& Vehicular behavior: Trajectories, vehicle speed, acceleration, turn recognition; Driver behavior: Turning intention,  aggression, perception reaction time; Pedestrian behavior: Motion prediction,  waiting time, walking speed, crossing speed, and choices; Safety assessment:  Gap analysis, threat,  risk,  conflict, accident;  Intersection safety systems: Driver assistance systems (driver perception enhancement, action suggestion and human driver interface, advanced vehicle motion control delegation), infrastructure-based systems (roadside warning systems, dilemma zone protection systems, decision support systems).	\\   
\hline     
Ahmed (2018) \cite{2018_SA_Ahmed} 
				& Trajectory-based analysis
							& Trajectory analysis: Datasets, extraction,  representation, applications; Clustering algorithms; Event detection: Methods and learning procedures; Localization of abnormal events: Methods and learning procedures; Video summarization and synopsis generation.\\     
\hline 
Lopez-Fuentes (2018) \cite{2018_L_Lopez-Fuentes} 
				& Emergency management using computer vision 
							& Emergency classification: Natural, human made (road accident, crowd related, weapon threat, drowning, injured person, falling person); Monitoring objective: Prevention, detection, response and understanding; Acquisition methods: Sensor location, sensor types, acquisition rate and sensor cost; Feature extraction algorithms: Color, shape and texture, temporal  (wavelet, optical flow, background modeling and subtraction, tracking) and convolution features;  Semantic information extraction using machine learning: Artificial neural networks, deep learning, support vector machines (SVMs), hidden markov models (HMMs), fuzzy logic.\\  
\hline
Mabrouk (2018)~\cite{2018_Mabrouk_abnormal}
				& Abnormal behavior recognition
							&Behavior representation; Anomalous behavior recognition methods: Modeling frameworks and classification methods, scene density and moving object interaction in crowded and uncrowded scenes; Performance evaluation: Datasets and metrics; Existing surveillance systems.\\								
\hline 
    \end{tabular}
\end{center}
\end{table*} 

Rest of the paper is organized as follows. 
First, the background and the terminologies used in the paper are introduced in Section~\ref{sec:Definitions}. Anomaly detection related visual scene learning methods are presented in Section~\ref{sec:Learning_methods}. Anomaly detection approaches and  classification are elaborated in Section~\ref{sec:Anomaly_Detection_Techniques}. Features used for anomaly detection and application areas are presented in Sections~\ref{sec:Features} and \ref{sec:Applied_areas}, respectively. A critical analysis of the existing methods followed by discussions on the challenges and future possibilities of anomaly detection are presented in Section~\ref{sec:Discussion}. We conclude the paper in Section~\ref{sec:Conclusion}.

\section{Computer Vision Guided Anomaly Detection Studies}
\label{sec:Survey}
\subsection{Background and Terminologies}
\label{sec:Definitions}
Features are assumed as data in the present context and are represented in the form of feature descriptors. Data typically occupy  a position in a multi-dimensional space depending on the feature descriptor length.

 Anomalies are data patterns that do not conform to a well-defined notion of normal behavior~\cite{2009_Chandola_anomaly}. There has been other synonyms of anomalies such as outliers, novelty in various application areas~\cite{2014_M_Gupta}. In this paper, we use  anomaly or outlier in the subsequent part.

\subsubsection{Anomaly Classification}
Traditionally, anomalies are classified as \textit{point} anomalies~\cite{2013_Roshtkhari_line, 2014_li_anomaly_Dynamic_Texture_Model, 2014_jeong_two_trajectory_hybridlearningLDA_GMM_probability_threshold_measure}, \textit{contextual} anomalies~\cite{2007_Song_conditional, 2015_yuan_online_feature_SCD_Anomalyby_EMDandMotiondifference} and \textit{collective} anomalies~\cite{2014_wang_detection_OCSVM, 2015_Cheng_Regression}. Data correspond to point anomaly if they are far away from the usual distribution. For example, a non-moving car on a busy road can be termed as a point anomaly.  Contextual anomalies correspond to data that may be termed normal in a different context. For example, in a slow moving traffic, if a biker rides faster as compared to others, we may term it as anomaly. Conversely, in a less dense road it may be a normal behavior. A group of data instances together may cause anomaly even though individually they may be normal. For example, a group of people dispersing within a short span of time can be termed as collective anomaly.

In the context of visual surveillance, it is common to see anomalies classified as \textit{local} and \textit{global} anomalies~\cite{2015_guo_Multicamera, 2016_Hu_Deep_Features, 2016_patil_global_HOFO_OCSVM, 2013_Yang_trasmil, 2015_pathak_Feature_HOG_HOF_pLSA_anomaly_projectionmodel_histogrambased, 2017_Sabokrou_DNN}. Global anomalies can be present in a frame or a segment of the video without specifying where exactly it has happened\cite{2015_guo_Multicamera, 2016_Hu_Deep_Features, 2016_patil_global_HOFO_OCSVM}. Local anomalies usually happen within in a specific area of the scene, but may be missed by global anomaly detection algorithms~\cite{2013_Yang_trasmil, 2015_pathak_Feature_HOG_HOF_pLSA_anomaly_projectionmodel_histogrambased, 2017_Sabokrou_DNN}. Some methods can detect both global and local anomalies\cite{2018_Wang_OCSVM_MotionDescriptor,  2016_abdallah_feature_3_Model_Hybrid, 2015_Cheng_Regression, 2018_Kaltsa_multiple_HDP, 2014_zhu_sparse_Histogram_of_Optical_Flow_Sparsecoding_reconstructioncost}.

\subsubsection{Challenges and Scope of Study}
The key challenges in anomaly detection are:  
(i) defining a representative normal region, (ii) boundaries between the normal and anomalous regions may not be crisp or well defined, (iii) the notion of anomaly is not same in all application contexts, (iv) limited availability of data for training and validation, (v) data is often noisy due to inaccurate sensing, and (vi) normal behavior evolves over time. 

We have done this survey based on the studies conducted on videos captured through a static camera. Anomaly detection using multiple cameras include additional challenges and the frameworks can be completely different~\cite{2015_Babaei_Multicamera_vehicles, 2015_guo_Multicamera}.

\subsection{Learning Methods}
\label{sec:Learning_methods}
Learning the normal behavior is not only relevant for anomaly detection, but also for diverse use cases. Pattern analysis~\cite{2014_Emonet_PatternAnalysis_Motif}, classification~\cite{2010_Nascimento_Classification}, prediction~\cite{Morris_2011_Prediction}, density estimation~\cite{2017_zhang_densityestimation}, and behavior analysis~\cite{2015_Bastani_BehaviorAnalysis} are a few amongst them.   

Learning methods can be classified as \textit{supervised}, \textit{unsupervised} or \textit{semi-supervised}. In supervised learning, the normal profile is built using labeled data~\cite{2000_Kamijo_Supervisedlearning, 2013_ji_3d_Supervisedlearning, 2014_karpathy_large_Supervisedlearning, 2014_simonyan_very_Supervisedlearning}. It is typically applied for classification and regression related applications. In unsupervised learning, normal profile is structured from the relationships between elements of the unlabeled dataset~\cite{2015_Srivastava_unsupervised}. Semi-supervised learning primarily uses unlabeled data with some supervision with a small amount of labeled data for specifying example classes known apriori~\cite{2018_sun_SemisupervisedLearning, 2017_liu_SemisupervisedLearning}. If learning happens through interactive labeling of data as and when the label info is available, such a learning is called active learning~\cite{2017_Varadarajan_semisupervised, 2017_de_active, 2018_liu_deep_active, 2015_nguyen_bayesian_feature_Optflowvector_active_iHMM_PCA_BN}. Such methods are  used  when unlabeled data are abundant and manual labeling is expensive. Reinforcement learning, a relatively new learning applied on computer vision, is an area of machine learning concerned with how software agents (discriminant and generator) ought to take actions in an environment so as to maximize some notion of cumulative reward~\cite{2018_wang_video_reinforcement, 2018_wang_reinforcement, 2017_zhang2_deep_reenforcement}. Some of the important works are summarized in Table~\ref{Tab:Learning_Categorization}.

\begin{table}[!h]
 \caption{Broad categorization based on learning methods}
\label{Tab:Learning_Categorization}
\begin{tabular}{p{0.13\textwidth} p{0.28\textwidth}}
\hline
 \textbf{Learning method}     &  \textbf{Ref.} \\ 
\hline
Supervised   & \cite{smeureanu2017deep, 2008_Piciarelli, 2018_DSingh_Autoencoder, 2013_CLiu_Sparse, 2017_Chong_Autoencoder, luo2017revisit_Sparse, 2016_hasan_Feature_handcrafterandAuto_Model_Autoencoder,2017_hinami_Feature_Auto_Model_FasterRCNN, 2000_Kamijo_Supervisedlearning, 2013_ji_3d_Supervisedlearning, 2014_karpathy_large_Supervisedlearning, 2014_simonyan_very_Supervisedlearning, 2016_zhou_spatial_supervised_Automaticfeatures_STC, 2016_chen_Supervised_Classifier_Feature_OS_SIFT_EML, 2018_lee_stan_Supervised_input_realandfakesequence_ModelGAN_Anomaly_deviation}\\ 
Unsupervised & \cite{farooq2017unsupervised, 2015_nGan_DPMM_outlier, xu2015learning_multifeature_SVM, ren2015unsupervised_Dictionary, javan2013online, 2016_Maiorano_Roughset, 2018_Liu_Relative_Density, 2013_Roshtkhari_line, 2015_Srivastava_unsupervised, 2017_vu_deep_Unsupervised, xiao2015learning_unsupervised, 2018_wei_unsupervised}\\
Semi-supervised & \cite{Wang2018_Semisupervised_Sparse, chakrabortyfreeway, ma2016density_semisupervised, 2018_sun_SemisupervisedLearning, 2017_liu_SemisupervisedLearning, 2015_pathak_Feature_HOG_HOF_pLSA_anomaly_projectionmodel_histogrambased}\\
\hline
\end{tabular}
\end{table}

 Learned models are not only been used in feature extraction, but also used in object detection~\cite{2014_Wang_Object_Detection}, classification~\cite{2014_karpathy_Classification}, activity recognition~\cite{2013_Nascimento_ActivityRecognition}, segmentation~\cite{2015_Khoreva_Segmentation}, tracking~\cite{2013_Wang_Object_Tracking_SparsePrototype}, entity re-identification~\cite{2015_liao_Reidentification}, object interaction analysis~\cite{2014_yu_interaction}, anomaly detection~\cite{2014_Kaltsa_STV}, etc.  Table~\ref{Tab:Learning_classification} presents some important learning methods used in anomaly detection.

\begin{table*}[!h]
 \caption{Examples of learning methods used in anomaly detection}
\label{Tab:Learning_classification}
\begin{tabular}{ p{0.1\textwidth} | p{0.2\textwidth} | p{0.61\textwidth} }
  \hline 
  \multicolumn{1}{l|}{\textbf{Learning Method}}    & \textbf{Method} & \textbf{Applied context}  \\
  \hline
  \multirow{2}*{Supervised}     &Hidden Markov Model (HMM)~\cite{1966_Baum_HMM} &  A supervised statistical Markov model where the system modeled is assumed to be a Markov process with hidden states: Used for anomaly detection in~\cite{2014_Biswas_HMM, 2018_wang_Modeling_HMM}. \\
  			\cline{2-3}
        					& Support Vector Machine (SVM)~\cite{1998_Hearst_SVM}  & A representation of data points in space, mapped such that separate categories are divided by a clear separation between them: A special class of SVM, namely One class SVM (OCSVM) has been extensively for anomaly detection~\cite{2001_Scholkopf_OCSVM}. \\
  			\cline{2-3}
        					& Gaussian Regression (GR)~\cite{2004_Rasmussen_GaussianRegression}  & A generic supervised learning method designed to solve regression and probabilistic classification problems: Used in~\cite{2015_Cheng_Regression, 2015_sabokrou_real_STC_GaussianClassifier} for anomaly detection from videos. \\  
  			\cline{2-3}
        					& Convolutional Neural Networks (CNN)~\cite{2016_goodfellow_CNN_Deep}  & A class of deep neural networks,  applied usually to analyze visual imagery: Due to its applicability in extracting semantic level features from the input, it has become popular in many applications including anomaly detection~\cite{2016_Hu_Deep_Features, 2016_medel_Feature_Auto_Model_LSTM}.\\           		  			\cline{2-3}
        					& Multiple Instance Learning (MIL)~\cite{babenko2009visual_MIL}  & A special learning framework which deals with uncertainty of instance labels: Instead of receiving a set of instances which are individually labeled, the learner receives a set of labeled bags, each containing many instances. If all the instances in it are negative, the bag may be labeled negative. If there is at least one positive instance, the bag is labeled positive. It has been used for anomaly detection in~\cite{2013_Yang_trasmil, sultani2018real_supervised}. 
        					\\  \cline{2-3}
        					& Long short-term memory (LSTM) networks~\cite{1997_Hochreiter_LSTM}  & A special kind of recurrent neural network (RNN) used in time series applications: In~\cite{luo2017revisit_Sparse, 2017_Luo_DNN, 2016_medel_Feature_Auto_Model_LSTM, 2015_Srivastava_unsupervised}, it has been used for anomaly detection. 
        					\\	\cline{2-3}
        					 
        					&  Fast Region-based-CNN (Fast R-CNN)~\cite{2015_Girshick_fasterRCNN}
        						& A higher variation of neural deep neural networks (DNN) that works efficiently in object classification over conventional CNNs:  Used for anomaly detection in~\cite{2017_hinami_Feature_Auto_Model_FasterRCNN}.\\ 					
  \hline  
  \multirow{3}*{Unsupervised}       &Latent Dirichlet Allocation (LDA)~\cite{2003_blei_method_LDA} & A topic model using statistical analysis to retrieve underlying topic distribution of in documents: Used for modeling visual words of videos for anomaly detection~\cite{2014_jeong_two_trajectory_hybridlearningLDA_GMM_probability_threshold_measure}.  \\
  			\cline{2-3}
        					& Probabilistic Latent Semantic Analysis (pLSA)~\cite{1999_hofmann_method_pLSA} & A model for representing co-occurrence  information  under  a  probabilistic framework: Used in~\cite{2015_kaviani_automatic_STCs_Hybrid_LDA_pLSA_FSTM} for anomaly detection. \\
  			\cline{2-3}
        					& Hierarchical Dirichlet Process (HDP)~\cite{2005_Teh_HDP} & A nonparametric Bayesian approach, built based on LDA, to cluster data: Used in data modeling and anomaly detection~\cite{2018_Kaltsa_multiple_HDP}. \\   
  			\cline{2-3}
        					& Gaussian Mixture Model (GMM)~\cite{2006_bengio_GMM} & A probabilistic model that assumes all the data points are generated from a mixture of a finite number of Gaussian distributions with unknown parameters: Used for anomaly detection in~\cite{Li2016_ImageDescriptor_GMM, 2017_wen_directional_feature_object_velocity_direction_GMM_anomaly_decisionmodel}.\\  
  			\cline{2-3}
        					& Density-based spatial clustering of applications with noise (DBSCAN)~\cite{1996_Ester_DBSCAN} & A density based non-parametric clustering algorithm used extensively for modeling and learning data patterns: Used for anomaly detection in~\cite{2015_ranjith_anomaly_Technique_Clustering}.\\           					          				  			\cline{2-3}
        					& Fisher kernel method~\cite{2010_perronnin_FisherLernel} & A function to measure similarity of two objects on the basis of sets of measurements for each object and a statistical model: Used to obtain trajectory feature representation in~\cite{2018_Wang_Semisupervised_SparseCoding}. \\   	
			\cline{2-3}
        					& Principal component analysis (PCA)~\cite{2011_jolliffe_PCA} & A statistical procedure of orthogonal transformation to convert a set of observations of possibly correlated variables into a set of values of linearly uncorrelated variables: Used for dimensionality reduction in~\cite{2018_wang_Adaboost_SVM_Hybrid}. \\   				
  			\cline{2-3}
        					& Particle Swarm Optimization~\cite{2011_kennedy_Motionmodel_PSO} & A  population based stochastic optimization technique: Used in~\cite{2014_Kaltsa_STV} to obtain optimized motion descriptor from a set of particles having individual motion characteristics.\\  
  			\cline{2-3}
        					& Generative Adversarial networks (GAN)~\cite{2014_Goodfellow_GAN} & A class of artificial intelligence algorithms used in unsupervised machine learning, implemented by a system of two neural networks (generator and discriminator) contesting with each other in a zero-sum game framework: Used for anomaly detection in~\cite{2017_Ravanbakhsh_Feature_Normal_OptFrames_Model_GAN_Anomaly_Reconstruction}.\\           					        					    
  \hline        					
  \multirow{2}*{Hybrid}     &HDP+HMM & A hybrid model: Used for representing sub-trajectories in~\cite{2013_Yang_trasmil} for anomaly detection using MIL. \\
  			\cline{2-3}
        					& GAN-LSTM~\cite{2018_lee_stan_Supervised_input_realandfakesequence_ModelGAN_Anomaly_deviation} & A hybrid model: Fake frames required for adversarial learning used in~\cite{2016_medel_ano_semisuper_ConvLSTM_predictedreconstructError} are generated using bidirectional Conv-LSTM~\cite{2015_xingjian_convLSTM}.\\   
  			\cline{2-3}
        					& CNN-LSTM~\cite{2016_medel_ano_semisuper_ConvLSTM_predictedreconstructError} & A hybrid model: Prediction-based anomaly detection with the help of CNN-LSTM.\\           					     					
  \hline 
\end{tabular}
\end{table*}
\subsection{Anomaly Detection Approaches}
\label{sec:Anomaly_Detection_Techniques}
\begin{figure*}[!h]
\forestset{%
  colour me out/.style={outer color=#1!75, inner color=#1!50, draw=darkgray, thick, blur shadow, rounded corners},
  rect/.append style={rectangle, rounded corners=2pt},
  dir tree switch/.style args={at #1}{%
    for tree={
      edge=-Latex,
      font=\sffamily,
      fit=rectangle,
    },
    where level=#1{
      for tree={
        folder,
        grow'=0,
      },
      delay={child anchor=north},
    }{},
    before typesetting nodes={
      for tree={
        content/.wrap value={\strut ##1},
      },
      if={isodd(n_children("!r"))}{
        for nodewalk/.wrap pgfmath arg={{fake=r,n=##1}{calign with current edge}}{int((n_children("!r")+1)/2)},
      }{},
    },
  },
}

\begin{forest}
  dir tree switch=at 1,
  for tree={
    font=\sffamily\bfseries,
    rect,
    align=center,
    edge+={thick, draw=darkgray},
    where level=0{%
      colour me out=green!50!white,
    }{%
      if level=1{%
        colour me out=magenta!50!orange!75!white,
        edge+={-Triangle},        
      }{%
        colour me out=magenta!50!orange!75!white,
        edge+={-Triangle},
      },
    },
  }
[Approaches
	[Model
    [Statistical
      [Parametric         
      ]
      [Non-parametric
      ]   
    ]
    ]
    [Proximity-based
        [Relative density]
        [Distance]      
    ]
    [Classification
      [SVM]
      [Bayesian]  
      [Rule-based]
    ]    
    [Reconstruction
    	[Sparse]
    	[PCA]
    	[Autoencoder]    
    ]      
    [Prediction
    ]             
    [Others
    	[Cluster]
    	[Fuzzy]
    	[Heuristic]
    	[Hybrid]
    ]           
    ]      
]
\end{forest}
\caption{Classification of the anomaly detection methods based on different approaches.}
\label{Fig:Approaches}
\end{figure*}
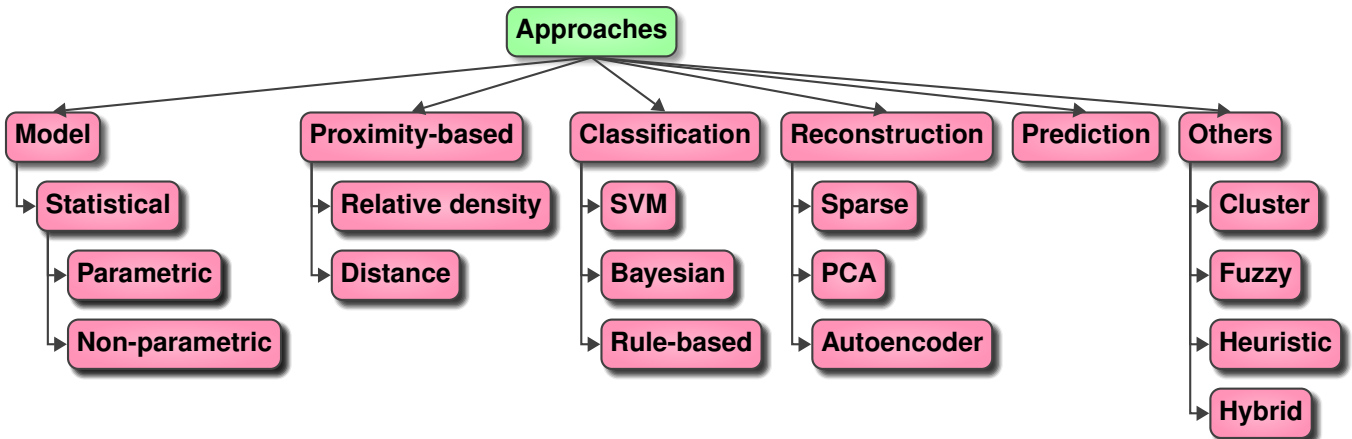

Anomaly detection approaches can be classified as depicted Fig.~\ref{Fig:Approaches}. 

\subsubsection{Model-based}
Model-based approaches learn the normal behavior of data by representing them  in terms of a set of parameters. Statistical approaches are used in general to learn the parameters of the model as they try to fit the data into a stochastic model. Statistical approaches may be either parametric or non-parametric. Parametric methods assume that the normal data is generated through parametric distribution and probability density function. Examples are Gaussian mixture models~\cite{Li2016_ImageDescriptor_GMM}, Regression models~\cite{2015_Cheng_Regression}, etc. In nonparametric statistical models, the  structure is not defined apriori, instead determined dynamically from the data. Examples are histogram-based~\cite{2016_zhang_feature_STG_OFM_Model_HOG_SVDD}, Dirichlet process mixture models (DPMM)~\cite{2015_nGan_DPMM_outlier}, Bayesian network-based models~\cite{2014_blair_event_Bayesean}, etc. Bayesian network estimates the posterior probability of observing a class label from a set of normal class labels and the anomaly class labels, given a test data instance. The class label with the biggest posterior is regarded as predicted class for the given test instance. Typically, topic model-based anomaly detection methods use Bayesian nonparametric approaches~\cite{2015_mousavi_analyzing,  2015_kaviani_automatic_STCs_Hybrid_LDA_pLSA_FSTM}. DNN-based models can also be categorized under parametric models, where the parameters are the weights and biases of the neural networks~\cite{2017_Sabokrou_DNN, chalapathy2017robust_DNN, 2017_Luo_DNN}. However, some researchers consider them as a classification approaches~\cite{li2016anomaly}, while many approaches (statistical, classification, information theoretic, reconstruction based) are used in the anomaly detection. Neural network-based  methods also adopt information theoretic approach to reduce cross entropy between expected and the predicted outputs in the model learning~\cite{AdamOptimizer}.  Hence, it may be also categorized under hybrid approaches.

\subsubsection{Proximity-based}
In proximity based approaches, anomalies are decided by how close they are to their neighbors. In distance-based approaches, the assumption is that normal data have dense neighborhood~\cite{2017_colque_histogramsofopticalflow_Informationtheoritic_entropy}. Density-based approaches compare the density around a point with the density around its local neighbors. The relative density of a point compared to its neighbors is computed as an outlier score~\cite{2018_Liu_Relative_Density}. 
\subsubsection{Classification-based}
Classification based anomaly detection methods assume that a classifier  can distinguish between normal and anomalous classes in a given feature space. Class-based anomaly detection techniques can be divided into two categories: one class and multi-class. Multi-class classification-based anomaly detection techniques assume that the training data contain labeled instances of normal and anomalous classes. A data point is assumed anomalous if it falls in the anomalous class~\cite{2016_chen_Supervised_Classifier_Feature_OS_SIFT_EML}. One-class classification (OCC)-based anomaly detection techniques assume that all training data have one label~\cite{2018_Wang_OCSVM_MotionDescriptor, 2014_wang_detection_OCSVM, 2016_patil_global_HOFO_OCSVM, 2017_xu_Feature_SDAE_AnomalyOCSVM}. Such techniques learn a discriminative boundary around the normal instances using a one-class classification algorithm. Support Vector Machines (SVMs) can be used for anomaly detection in the one-class setting extensively in visual surveillance~\cite{2009_Chandola_anomaly, 2016_patil_global_HOFO_OCSVM}.  Rule-based approaches learn rules that capture the normal behavior of a system~\cite{2012_Saligrama_rule_based}. A test instance that is not covered by any such rule, is considered as an anomaly. 
\subsubsection{Prediction-based}
Prediction-based approaches detect anomaly by calculating the variation between predicted and actual spatio-temporal characteristics of the feature descriptor~\cite{2017_Liu_future}. HMM and LSTM models rely on such approaches for anomaly detection~\cite{2014_Biswas_HMM, 2016_medel_Feature_Auto_Model_LSTM, 2016_medel_ano_semisuper_ConvLSTM_predictedreconstructError}.
\subsubsection{Reconstruction-based}
In reconstruction-based techniques, the assumption is, normal data can be embedded into a lower dimensional subspace in which normal instances and anomalies appear differently. Anomaly is measured based on the data reconstruction error. Some of the examples are, sparse coding~\cite{ tan2016fast_Sparsecode, zhang2016abnormal_Sparsecode, 2017_yu_content_Sparse},  autoencoder~\cite{2016_hasan_Feature_handcrafterandAuto_Model_Autoencoder}, and principal component analysis (PCA)-based approaches~\cite{2018_Liu_Relative_Density}. 
\subsubsection{Other Approaches}
There are two types of clustering approaches. One relies on an assumption that the normal data lie in a cluster, while anomaly data do not get associated with any cluster~\cite{2015_ranjith_anomaly_Technique_Clustering}. The later type is based on an assumption that normal data instances belongs to big and dense clusters, while anomalies either belong to little/small clusters. Fuzzy inference systems take a fuzzy data point and uses the rules related to membership and strength at which data point fires the rules to decide whether the data is anomalous or not~\cite{wijayasekara2014mining, 2018_li_Fuzzy_road}. Heuristic methods intuitively decide about the feature values, spatial location, and contextual information to decide on anomalies. However, many practical systems do not entirely depend on one technology, rather hybrid approaches are used for  anomaly detection~\cite{2018_wang_Adaboost_SVM_Hybrid, chen2015detecting_Hybrid_technique, 2014_mo_adaptive_Sparse_hybrid}. Table~\ref{Tab:Specific_Anomaly_Techniques} presents the aforementioned categorization.
\begin{table}[!h]
 \caption{Specific categorization of anomaly detection techniques}
\label{Tab:Specific_Anomaly_Techniques}
\begin{tabular}{p{0.15\textwidth} p{0.29\textwidth}}
\hline
 \textbf{Specific Techniques}     &  \textbf{Ref.} \\ 
\hline
SVM   				& \cite{smeureanu2017deep,2008_Piciarelli,xu2015learning_multifeature_SVM, 2018_Wang_OCSVM_MotionDescriptor, 2013_Isaloo_Semisupervised_SVM_Trajectory, 2014_batapati_video_SVM_TRAJECTORY, 2017singh_feature_STIP_graphs_Anomaly_SVM}\\ 
Sparse & \cite{2017_yu_content_Sparse, 2014_Biswas_Sparse, 2013_CLiu_Sparse, luo2017revisit_Sparse, Wang2018_Semisupervised_Sparse, ren2015unsupervised_Dictionary}\\
PCA & \cite{2014_li_anomaly_Dynamic_Texture_Model}\\
Autoenoder & \cite{2018_DSingh_Autoencoder, 2016_hasan_Feature_handcrafterandAuto_Model_Autoencoder, 					2017_Chong_Autoencoder, 2018_Ribeiro_Autoencoder}\\
Regression			&\cite{2018_wang_Adaboost_SVM_Hybrid,2015_sabokrou_real_STC_GaussianClassifier}\\
Density-based 		&\cite{farooq2017unsupervised, ma2016density_semisupervised}\\
Clustering-based	&\cite{2016_ghrab_abnormal_trajectoryfeature, 2015_nGan_DPMM_outlier,2017_Varadarajan_semisupervised}\\ 
Statistical methods	&\cite{javan2013online, 2013_Chockalingam_AVSS, Li2016_ImageDescriptor_GMM}\\
Prediction					&\cite{2014_Biswas_HMM, 2016_medel_Feature_Auto_Model_LSTM, 2015_Kruthiventi_dominant_CRF_Clustering_distance, 2014_akoz_traffic_HMM_SVM_Nearest_neibour}\\
Bayesian Networks	&\cite{isupova2016anomaly}\\
Fuzzy logic-based 	& \cite{wijayasekara2014mining, 2018_li_Fuzzy_road}\\
Hybrid				&\cite{2013_Yang_trasmil, 2018_Ribeiro_Autoencoder, 2018_Liu_Relative_Density, 2018_Xu_dual_Hybrid, 2014_mo_adaptive_Sparse_hybrid, 2017_hinami_Feature_Auto_Model_FasterRCNN,  2015_ullah_traffic_hybrid_Thermodynamics, 2016_abdallah_feature_3_Model_Hybrid, 2014_multi_hu_Trajectory_hrybrid_Motionpatternlearning_anymaybymotionpatternmatching}\\
Relative Density	&\cite{2018_Liu_Relative_Density}\\
Heuristic			&\cite{2018_wei_unsupervised, 									 2018_Chang_video_Heuristic, 2014_Yun_motion_MIF_Heuristic,  2014_lee_hierarchical_Heuristic_hybrid, 2017_sukanyathara_motion_features_track_Heuristic, 2016_Maiorano_Roughset, tudor2017unmasking}\\
\hline
\end{tabular}
\end{table}
\subsection{Features Used in Anomaly Detection}
\label{sec:Features}
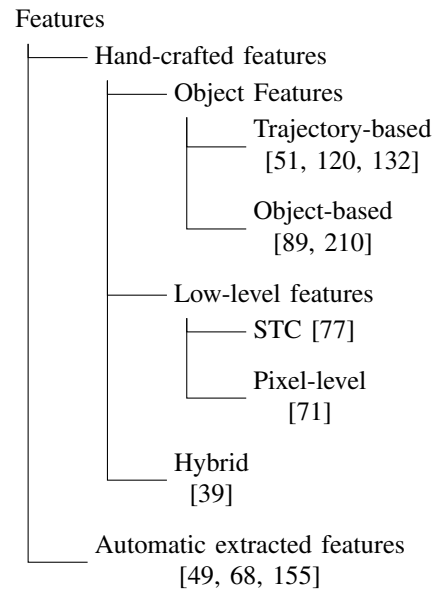
\begin{figure}[!h]
\centering    
\forestset{
  my tier/.style={
    tier/.wrap pgfmath arg={level##1}{level()},
  },
}                        
\begin{forest}
for tree={
    grow'=0,
    child anchor=west,
    parent anchor=south,
    anchor=west,
    calign=first,
    s sep+=-5pt,
    inner sep=2.5pt,
    edge path={
      \noexpand\path [draw, \forestoption{edge}]
      (!u.south west) +(7.5pt,0) |- (.child anchor)\forestoption{edge label};
    },
    before typesetting nodes={
      if n=1{
        insert before={[, phantom, my tier]},
      }{},
    },
    my tier,
    fit=band,
    before computing xy={
      l=30pt,
    },
  }
  [Features
	[Hand-crafted features
	    [Object Features
     		[Trajectory-based\\
     		\cite{2016_ghrab_abnormal_trajectoryfeature, 2017_trajectory_Heuristic, 2016_nghia_Feature_Trajectory_Surroundng}
     		]
     		[Object-based\\
     		\cite{2017_kumar_vehicle_object, 2015_yuan_online_feature_SCD_Anomalyby_EMDandMotiondifference}
     		]
    	]
    	[Low-level features
      		[STC\cite{2014_Kaltsa_STV}
      		]
      		[Pixel-level\\
      		\cite{isupova2016anomaly}    
      		]	
    	]
    	[Hybrid\\
    	\cite{2017_Cosar_Pixel_trajectory_Unsupervised}
    	]
	]
    [Automatic extracted features\\
    \cite{2016_Hu_Deep_Features, 2016_fang_Feature_MHOF_SI_PCANet, 2018_sabokrou_Feature_CNN_Anomaly_GaussianClassifier}
    ]
  ]
\end{forest}   
\caption{Overall classification of features used in anomaly detection.}
\label{Fig:Feature_taxonomy}
\end{figure}  
\begin{table*}[!h]
 \caption{Representative work based on used features}
\label{Tab:Feature_Representative_Categorization}
\scriptsize
\begin{tabular}{p{0.1\textwidth}p{0.1\textwidth} p{0.1\textwidth}p{0.3\textwidth}p{0.3\textwidth}}
\hline
\textbf{Ref.} & \textbf{Features}  & \textbf{Learning} &  \textbf{Anomaly Criteria} & \textbf{Highlight}\\ 
\hline
Yang (2013)~\cite{2013_Yang_trasmil}
		& Sub-trajectories
	& Multi instance learning 		
			& Nearest neighborhood based approach with Hausdorff distance-based threshold for anomaly detection.
				& Sub-trajectories-based local anomaly detection capability.\\
\hline
Roshtkhari (2013)~\cite{2013_Roshtkhari_line}
		& 3D Spatio-temporal volume
	& Code-book model		
			& Threshold applied on likelihood/saliency map. 
				& Fast anomaly localization requiring less training data. Does not require any feature analysis, background/foreground segmentation and tracking, and can be applied for real-time applications. \\
\hline
Li (2014) \cite{2014_li_anomaly_Dynamic_Texture_Model}
	& MDTs from Spatio-temporal patches  
		& Dynamic Texture Model   
			& Threshold on negative log-likelihood on temporal mixture of dynamic textures for temporal anomaly and threshold on the saliency for spatial anomalies.	
				& Detection of both temporal and spatial anomaly detection capability complex crowded scene.\\ 
\hline
Kaltsa (2014)~\cite{2014_Kaltsa_STV}
	&HOSA+HOGs over image patches
		& SVM
			&OCSVM based anomaly detection.
				& Robustness to local noise and anomaly detection detection in crowded scene.\\
\hline				
~Jeong (2014)\cite{2014_jeong_two_trajectory_hybridlearningLDA_GMM_probability_threshold_measure}	 	& Trajectories and pixel velocities
			& Hybrid (LDA + GMM)
				& Threshold on the probability score.
					& Thorough study conducted on at intersections and roads for traffic pattern analysis.	\\
\hline					
Zhu (2014)~\cite{2014_zhu_sparse_Histogram_of_Optical_Flow_Sparsecoding_reconstructioncost}	&Histogram of optical flow features
		& Sparse coding
			& Threshold on reconstruction cost used as anomaly measure.
				& The method can detect both local and global anomalies. Experiments though not conducted on traffic junctions though could be suitable for busy junctions.	\\
\hline					
Kaltsa (2015)~\cite{2015_kaltsa_swarm_Feature_HOG_HOS_PSO_Anomalyby_SVDD}		
	& Hybrid (HOS + HOG + PSO)
		& SVM
			& Support Vector Data Description (SVDD) method~\cite{tax2004support} for anomaly detection.
				&Swarm intelligence is exploited for the extraction of robust motion and  appearance features to model and to detect anomalies.\\	
\hline
Maousavi (2015)~\cite{2015_mousavi_analyzing}
	&Histogram of Oriented Tracklets (HOT)
		&LDA
			& Log-likelihood based fixed threshold of visual words for anomaly detection.
				& Comprehensive evaluation using topic model based anomaly detection and localization for a wide range of real-world videos.\\
\hline
Cheng (2015)~\cite{2015_Cheng_Regression}
	&Spatio-temporal interest points (STIPs)~\cite{2005_dollar_Feature_STIP}
		&Gaussian regression
			& Local anomalies: k-NN-based likelihood threshold with respect to the visual vocabulary of STIP codebook. Global anomalies: Using global negative log likelihood threshold.
				&  STIPS effectively used for local and global anomaly detection.\\
\hline

Mendel (2016)~\cite{2016_medel_Feature_Auto_Model_LSTM}
	& Automatic videos features with CNN.	
		&Conv-LSTM
			&Reconstruction error between predicted and actual output.
				& Effective for recognizing abnormalities when the training data is loosely supervised to contain mostly normal events.\\
\hline
Zhang (2016)~\cite{2016_zhang_video_feature_opticalflow_histogram_STC_method_heuristic_LSH_anomaly_PSO}		& Histogram of optical flow
			& Clustering
				& Anomaly score based on Hamming distance.
					& Locality sensitive hashing filters used in anomaly detection.\\ 
\hline					
Lan (2016)~\cite{2016_lan_real_Feature_HOG_Heuristic_hybrid_anomalyusing_relativemotion}		&HOG
			& Heuristic method 
					& Anomalies detected using relative speeds of detected objects.
					&An interesting study about abandoned objects that could possibly cause traffic accidents or some other untoward incidents.\\
\hline						
Hasan (2016)~\cite{2016_hasan_Feature_handcrafterandAuto_Model_Autoencoder}
	& Handcrafted HOG+HOF~\cite{2013_wang_Feature_HOG_HOF} and automatic CNN extracted features
		&Dual Autoencoder model
			& Anomaly score, namely regularity score derived using reconstruction error in autoencoders.
				&  A regularity score, used as a measure of normalcy in a scene, derived using both hand crafted features and automatic features using fully convolutional feed-forward
autoencoder. \\	
\hline				
Hinami (2017)\cite{2017_hinami_Feature_Auto_Model_FasterRCNN}
	&Deep features from CNN
		& Multi-test Fast R-CNN.
			& Anomaly detection with a combination of semantic features using (a)Nearest neighbor-based method (NN), (b)OCSVM and (c) KDE. 
				& It addresses the problem of joint detection and
recounting of abnormal events in videos in presence of false alarms.\\					
\hline					
Wen (2017)~\cite{2017_wen_directional_feature_object_velocity_direction_GMM_anomaly_decisionmodel}		& Object (velocity and direction)
			& GMM
				& Model based anomaly detection.
					& Speeding events detection that could be relevant on road, though authors have tested the method for indoor scenarios.\\
\hline
Ravanbakhsh (2017)~\cite{2017_Ravanbakhsh_Feature_Normal_OptFrames_Model_GAN_Anomaly_Reconstruction}
	&Opticalflow frames + Normal frames
		&GAN
			&Anomaly score as a fusion of Optical-flow and appearance reconstruction error. 
				& Global and Local anomaly detection in crowded scene.\\
\hline	
Lin (2017)~\cite{2017_lin_tube_trajectorybased}
		& 3D-Tube
	& SVM		
			& Contextual information embedded in trajectory thermal transfer fields using OCSVM.
				& This is first kind of anomaly detection done using thermal fields that can detect contextual anomalies.\\
\hline				
Liu (2017)~\cite{2017_Liu_future}
		& Automatically extracted optical flow, intensity and gradient features.
	& GAN		
			& Peak Signal to Noise Ratio (PSNR) score based on optical flow, intensity, gradient loss. 
				&  DNN-based prediction (\cite{2015_ronneberger_uNet_Prediction}) and GAN~\cite{2016_Goodfellow_nips} based discriminator applied on optical flow frames derived using (\cite{2015_Dosovitskiy_flownet}) to detect robustness to the uncertainty in normal events and the sensitivity to abnormal events.\\
\hline				
Colque (2017)~\cite{2017_colque_histogramsofopticalflow_Informationtheoritic_entropy}
	& HOFME
		&  Histogram based model
			&  Nearest Neighbor threshold.
				& A new feature descriptor HOFME that could handle diverse anomaly scenarios as compared with conventional features. \\
\hline					
Giannakeris (2018)~\cite{2018_Giannakeris_speed}
		& Trajectory Fisher vector
	& SVM		
				&  Anomaly score derived from the Fisher vector using OCSVM. 
					& Anomaly detection done using robust optical flow descriptors of the detected vehicles with the use of DNNs
and Fisher vector representations from spatiotemporal visual
volumes.\\	
\hline					
Lee (2018)~\cite{2018_lee_stan_Supervised_input_realandfakesequence_ModelGAN_Anomaly_deviation}
	&Real and Fake frames 
		&GAN
			& Abnormality score derived using the losses of the generator and the discriminator. 
				& Can detect anomalies from dataset containing complex motion and frequent occlusions.\\
\hline				
Kalta (2018)~\cite{2018_Kaltsa_multiple_HDP}
		& Code words of spatio-temporal regions 
	& Multiple HDPs		
				& Confidence score of reconstruction of region clips.
					&  Both local and global anomaly detection using super-pixels and interest point tracking~\cite{2012_Achanta_slic_Superpixel} applied on real-life videos.\\
\hline					
Sultani (2018)\cite{sultani2018real_supervised}
	&  Video clips
		&   Deep MIL Ranking Model
			& An anomaly score using sparsity and smoothness constraints.
				& A generic method applied on a variety of real-life scenarios. \\									
\hline
\end{tabular}
\end{table*}
\begin{table*}[!h]
 \caption{Representative work on scope of applied areas}
\label{Tab:Representative_Application}
\scriptsize
\begin{tabular}{p{0.1\textwidth}p{0.15\textwidth} p{0.1\textwidth}p{0.35\textwidth}p{0.18\textwidth}}
\hline
\textbf{Ref.}      & \textbf{Technique} & \textbf{Scene} & \textbf{Anomalies} & \textbf{Datasets}\\ 
\hline
Yang (2013)~\cite{2013_Yang_trasmil}
	& Multi instance learning 
		&Lobby.
			& One person walking, browsing, resting, slumping or fainting, leaving bags behind, people/groups meeting, walking together and then splitting up and two people fighting.
				&CAVIAR.\\
\hline				
Roshtkhari (2013)~\cite{2013_Roshtkhari_line}
	& Code-book (Sparse) model
		& Subway, walkway.
			& Abnormal walking patterns, crawling, jumping over objects, falling down, non-pedestrians on a walkway, walking in the wrong direction, irregular interactions between people and some other events including sudden stopping, running fast, walking in the wrong direction and loitering.
				&UCSC (Ped1, Ped2), Bellview and Person. \\	
\hline						
Jeong (2014)~\cite{2014_jeong_two_trajectory_hybridlearningLDA_GMM_probability_threshold_measure}
	& LDA + GMM
		& Junctions, walkway, roads, public gathering area.
			& Illegal U-turn, vehicle in opposite direction, disordering in the the traffic signal, over speed on a pavement, unusual crowds speed, a car stops on a railway.
				&UCSC, UMN, MIT, QMUL and In-house datasets.\\
\hline				
Li (2014)~\cite{2014_li_anomaly_Dynamic_Texture_Model}				
	& Dynamic Texture model
		&Walkways, junction.
			& Non-pedestrian entities in the walkways, people walking across a walkway or in the surrounding grass, U-turn.
				& UCSD (Ped1, Ped2), U-turn and UMN.\\
\hline				
Mo (2014)~\cite{2014_mo_adaptive_Sparse_hybrid}
	&Sparsity Model + OCSVM
		& Junction, road, parking lot.
			&Man suddenly falls on floor, vehicle almost hits a pedestrian, car violates the stop sign rule, car fails to yield to oncoming car while turning left, driver backs his car in front of stop sign.
				& i-LIDS, CAVIAR and In-house dataset namely XEROX.\\
\hline				
Patino (2014)~\cite{2014_Patino_multiresolution}
	&Statistical with heuristic approach	
			&Parking lot, road intersection.
					& Unusual object trajectories such as U-turn, vehicle stopping at pedestrian way, person stopping between two lanes outside zebra passages, person crossing lanes outside zebra passages, loitering and vehicle/person staying at a place for longer duration.
						& ARENA, CAVIAR and MIT trajectory dataset.\\
\hline						
Akos (2014)~\cite{2014_akoz_traffic_HMM_SVM_Nearest_neibour}
	& Hybrid (HMM + SVM + k-NN)
		&Intersection.
			&Collision, nearby passes.
				& NGSIM and AIRS.\\
\hline				
Wang (2014)~\cite{2014_wang_detection_OCSVM}
	&OCSVM
		& Walkway, public gathering place.
			& Local dispersion of crowds.
				&	PETS2009 and UMN.\\
\hline				
Yun (2014)~\cite{2014_Yun_motion_MIF_Heuristic}
	& Motion interaction field (MIF) symmetry model  
		& Junction.
			& Accident detection.
				& Car accident.\\
\hline				
Xia (2015)~\cite{2015_Xia_vision}
	& Low rank approximation on motion matrix created using optical flows.
			& Road, intersection.
				& Accident detection.
					&In-house dataset.\\
\hline						
Cheng (2015)~\cite{2015_Cheng_Regression}
	& Gaussian regression
			& Road, walkways, subway, intersection.
				& Non pedestrians appearing in walkway, chase, fight, run together, traffic interruption, jaywalk, illegal u-turn, strange driving.
					&UCSD (Ped1), Behave and QMUL.\\
\hline					
Xu (2015)~\cite{xu2015learning_multifeature_SVM}
	& Hybrid (DNN + Autoencoder + OCSVM)
		&Walkways.
			&Non pedestrians appearing in walkway.
				& UCSD(Ped1, Ped2).\\
\hline				
Kaviani(2015)~\cite{2015_kaviani_automatic_STCs_Hybrid_LDA_pLSA_FSTM}	
	& Hybrid (LDA+STC+pLSA+FSTM)
		&Roadways, Junctions.
			&Accident detection.
				& QMUL and In-house datasets.\\
\hline				
			
Nguyen (2015)~\cite{2015_nguyen_bayesian_feature_Optflowvector_active_iHMM_PCA_BN}	
	& Bayesiean non-parametric 
		&Junctions.
			&Street fight, loitering, truck-unusual stopping, big truck blocking camera.
				& MIT.\\
\hline				

Pathak (2015)\cite{2015_pathak_Feature_HOG_HOF_pLSA_anomaly_projectionmodel_histogrambased}
	&pLSA
		&Junction, highway, roadways.
			& Car stops after the stop-line, jaywalk, vehicle abruptly crossing the road.
				& ldiap, highway (In-house) and i-LIDS.\\
\hline				
Medel (2015)~\cite{2016_medel_ano_semisuper_ConvLSTM_predictedreconstructError}
	& ConvLSTM
		& Walkways, roadways.
			& People walking perpendicular.
to the walkway, or off the walkway, movement of non-pedestrian entities and anomalous pedestrian motions, pedestrians walking off the walkway.
				& USCD (Ped1, Ped2) and Avenue.\\
\hline				

Zhou (2016)~\cite{2016_zhou_spatial_supervised_Automaticfeatures_STC}
	& CNN
		& Junction, walkways, dispersing crowd.
			& U-turn, unexpected presence of vehicles.
				& UCSD, UMN, and U-turn.\\
\hline				
Zhang (2016)~\cite{2016_zhang_feature_STG_OFM_Model_HOG_SVDD}
	& Hybrid (Histogram of Optical flow and Support Vector Data Description)
		&Walkways.
			& Non pedestrians on walkways.
				& UCSD ped1.\\
\hline				

Xu (2017)~\cite{2017_xu_Feature_SDAE_AnomalyOCSVM}
	& OCSVM with SDAE features
		& Walkways.
			& Non pedestrians on walkways.
				& UCSD.\\
\hline				
Vishnu (2017)~\cite{2017_vishnu_feature_HFG_anomaly_hybrid_MLR_DNN_congestion_oncount_threshold}
	& Hybrid (MLR+DNN+vehiclecount)
		& Highway, Roadway, Junction.
		& Congestion detection, ambulance detection, accident detection.
			& In-house datasets.\\
\hline			
Liu (2017)~\cite{2017_Liu_future}
	& Heuristic
			& Roadways, walkways, junction.
				& Throwing objects, loitering and running, non pedestrians on walkways, presence of people at unexpected area of road.
					&Avenue, UCSD Ped1, UCSD Ped2 and ShanghaiTech.\\	\hline		
Giannakeris (2018)~\cite{2018_Giannakeris_speed}
	& SVM
			& Roadways.
				& Car crashes, stalled vehicles.
					&NVDIA CITY.\\	
\hline					
Chebiyyam (2017)~\cite{2017_chebiyyam_Feature_RAG_trajectoryFeatureVector_Anomaly_SVM}
 	& Heuristic using SVM and Region Association Graph
 		& Parking lot, walkways. 
 			& Object encircling a particular regions, target switching between two or more regions for a sustained period of time.
 				& MIT Parking trajectory, Avenue and a Custom dataset.\\
\hline 				
Yun (2017)~\cite{2017_yun_Feature_MIF_Anomaly_Sparse}
	& Sparse learning using motion interaction field ~\cite{2014_Yun_motion_MIF_Heuristic}
		& Junction, roadways, public gathering area.
			& Car accidents, crowd riots, and uncontrolled fighting.
				& BEHAVE, UMN and Car accident.\\	
\hline						
Wang (2018) \cite{2018_Wang_Semisupervised_SparseCoding}
	&Sparse topic Model
		& Junction, Roadways.
			& Car deviating from normal Pattern, Conflicting patterns, Vehicle suddenly interrupting normal pattern,  jaywalk, vehicle retrograde, pedestrian near collisions with vehicle.
				& i-LIDS and QMUL. \\
\hline
Kalta (2018)~\cite{2018_Kaltsa_multiple_HDP}
	& HDP
			& Intersections.
				& Jay walking, illegal U-turns, wrong vehicle direction, traffic break.
					&QMUL, ldiap and U-turn.\\	
\hline					
Sultani (2018)\cite{sultani2018real_supervised} 
	& Deep MIL Ranking Model 
		&   Intersection, roadways, walkways.
			& Abuse, arrest, arson, assault, accident, burglary, fighting, robbery.
				& UMN, UCSC (Ped1, Ped2), Avenue, Subway, BOSS, Ab
normal Crowd, and a set of Local datasets.	\\						
\hline
\end{tabular}
\end{table*} 
As mentioned earlier, anomaly detection is essentially done by applying specific technique on the extracted feature. However, in visual surveillance, primary data is a video which is a sequence of frames. Hence, it is essential to extract the relevant features from the videos as these features become input to the specific technique used in anomaly detection. The choice of feature plays a key role in the capability of detecting specific anomalies. In some methods, preprocessing essentially involves extracting the foreground information and applying specific techniques for finding objects from the foreground~\cite{2016_lan_real_Feature_HOG_Heuristic_hybrid_anomalyusing_relativemotion, 2014_li_anomaly_Dynamic_Texture_Model, 2015_ullah_traffic_hybrid_Thermodynamics, 2018_wei_unsupervised}. Also, histograms extracted from the pixel level features can become inputs to  anomaly detection methods~\cite{2014_wang_detection_OCSVM, 2015_wang_detection_Opticalflow_HOFO_SVM_PCA, 2016_zhang_video_feature_opticalflow_histogram_STC_method_heuristic_LSH_anomaly_PSO, 2017_colque_histogramsofopticalflow_Informationtheoritic_entropy}. Some methods use detected objects or object trajectories as inputs to the anomaly detection methods~\cite{2016_ghrab_abnormal_trajectoryfeature, 2017_lin_tube_trajectorybased, 2015_zhou_unusual}. Deep neural networks (DNN) extract features automatically and used them for anomaly detection~\cite{2017_vu_deep_Unsupervised, 2018_sabokrou_Feature_CNN_Anomaly_GaussianClassifier, 2018_lee_stan_Supervised_input_realandfakesequence_ModelGAN_Anomaly_deviation}.      

Feature are typically in the form of vectors, corresponding to the data. The method proposed in~\cite{2016_hasan_Feature_handcrafterandAuto_Model_Autoencoder} uses histograms of oriented gradients (HOG), histograms of optical flows (HOF), improved trajectory features~\cite{2013_wang_Feature_HOG_HOF}, and automatic features extracted using DNN. A mixture of dynamic textures has been used in~\cite{2014_li_anomaly_Dynamic_Texture_Model}. Histograms of oriented swarm accelerations (HOSA) coupled with histograms of oriented gradients (HOGs) has been used in learning~\cite{2014_Kaltsa_STV}. Authors of~\cite{2017_lin_tube_trajectorybased} have used 3D-tube representation of trajectories as features using the contextual proximity of neighboring trajectory for learning normal trajectory. In~\cite{2018_Giannakeris_speed}, Fisher vector corresponding to each trajectory obtained using optical flow of the object and its position, has been used. Histogram of optical flow and motion entropy (HOFME) have been used in~\cite{2017_colque_histogramsofopticalflow_Informationtheoritic_entropy}. In DNN-based systems, high level features are automatically extracted. 

Broadly, the features can be classified as object oriented and non-object oriented. The classification is represented in Fig.~\ref{Fig:Feature_taxonomy}. Using object oriented features, anomalies can be detected by extracting the objects~\cite{2014_lim_isurveillance_objectfeature, 2017_kumar_vehicle_object} or  trajectories~\cite{2017_lin_tube_trajectorybased, 2016_ghrab_abnormal_trajectoryfeature, 2017_mehboob_trajectory_feature}. Objects or trajectories represented in the form of feature descriptors become the data for anomaly detection. In the latter approach, low-level descriptors for pixel or pixel group features, intensities, optical flows, or resultant features from spatio-temporal cubes (STC)~\cite{2014_li_rapid_lowlevelfeature, 2014_nallaivarothayan_mrf_stc, 2014_kaviani_incorporating_stc, 2016_traffic_STC, 2018_zhou_anomaly_Opticalflow_features, 2014_rasheed_tracking_Opticalflow_neuralnetwork, 2015_sabokrou_real_STC_GaussianClassifier} have been used for anomaly detection. Some methods use hybrid features for anomaly detections~\cite{2013_li_abnormal_hybrid_features, 2013_kwon_scene_hybrid_features_and_methods, 2014_elahi_computer_object_and_motion_features, 2017_Cosar_Pixel_trajectory_Unsupervised}. Some of the important work using various aforementioned features are summarized in Table \ref{Tab:Feature_Representative_Categorization}.
\subsection{Applied Areas}
\label{sec:Applied_areas}
In this section, we discuss the research work that have been carried out so far focusing on scene and datasets. Typical scenes are road segments, junctions, parking areas, highways, pedestrian paths, etc.  A few of the important research work have been summarized in Table~\ref{Tab:Representative_Application}. We mainly highlight the underlying  techniques,  applicable scenes, anomaly types and datasets. The datasets  often used in such work are QMUL~\cite{2012_hospedales_video_QMUL_dataset}, CAVIAR~\cite{2004_CAVIAR_dataset}, UCSD~\cite{2010_Mahadevan_UCSCPed},  Bellview~\cite{2010_Zaharescu_Bellview}, Person~\cite{2008_Adam_Persondataset}, UMN~\cite{2009_Mehran_UMN_dataset}, ARENA~\cite{2014_Patino_multiresolution}[check again], Avenue~\cite{2016_Hasan_Avenue_dataset}, U-turn~\cite{2009_Benezeth_abnormal_Uturn_dataset}, MIT Trajectory~\cite{2010_wang_MIT_Trajdataset}, MIT~\cite{2009_Wang_MIT_dataset}, MIT parking trajectory~\cite{2008_Wang_MIT_Parking_trajectory}, NGSIM~\cite{NGSIM_dataset}, AIRS~\cite{AIRS_dataset}, PETS2009~\cite{2009_Ellis_pets2009}, Behave~\cite{2010_Blunsden_Behave_dataset}, i-LIDS~\cite{iLIDS_dataset},  ShanghaiTech~\cite{luo2017revisit_Sparse}, NVDIA CITY~\cite{CITY_CHALLENGE}, BOSS~\cite{sultani2018real_supervised}, Car Accident~\cite{2010_Sultani_abnormal_car_accident_dataset}, and ldiap~\cite{2009_topic_ldiap_dataset}. 
  
\subsection{Online vs. Offline}
Majority of the techniques applied for anomaly detection focus on online usage~\cite{2013_Roshtkhari_line, 2016_lan_real_Feature_HOG_Heuristic_hybrid_anomalyusing_relativemotion, 2014_blair_event_Bayesean, 2015_sabokrou_real_STC_GaussianClassifier, 2017_sukanyathara_motion_features_track_Heuristic, Morris_2011_Prediction, 2008_Adam_Persondataset}. Some methods~\cite{2013_CLiu_Sparse, 2014_kaviani_incorporating_stc,  2015_ranjith_anomaly_Technique_Clustering, 2015_kaviani_automatic_STCs_Hybrid_LDA_pLSA_FSTM} can be  termed near real-time because the detection can happen only by segmenting test videos from the real scene. Offline methods are also used in road networks though the results are not immediate especially for data analysis~\cite{2017_lin_tube_trajectorybased,  2018_Chang_video_Heuristic, 2016_Maiorano_Roughset}. However, online methods are more preferred since they generate instantaneous results. A categorization is presented in Table~\ref{Tab:Application_Type}.
\begin{table}[!h]
 \caption{Online vs. Offline}
\label{Tab:Application_Type}
\begin{tabular}{p{0.15\textwidth} p{0.29\textwidth}}
\hline
 \textbf{Type}     &  \textbf{Ref.} \\ 
\hline
Online   				& \cite{Wang2018_Semisupervised_Sparse,2018_Wang_OCSVM_MotionDescriptor, 2018_Ribeiro_Autoencoder, 2018_wang_Adaboost_SVM_Hybrid, 2018_Liu_Relative_Density, 2018_wei_unsupervised, 2018_Giannakeris_speed, 2018_Xu_dual_Hybrid, 2017_Liu_future, 2018_sabokrou_Feature_CNN_Anomaly_GaussianClassifier, 2017_Varadarajan_semisupervised, 2017_yu_content_Sparse, 2017_vu_deep_Unsupervised, 2017_wen_directional_feature_object_velocity_direction_GMM_anomaly_decisionmodel, 2017_sukanyathara_motion_features_track_Heuristic, 2017_colque_histogramsofopticalflow_Informationtheoritic_entropy, 2017_chebiyyam_Feature_RAG_trajectoryFeatureVector_Anomaly_SVM, 2017_trajectory_Heuristic, 2017_vishnu_feature_HFG_anomaly_hybrid_MLR_DNN_congestion_oncount_threshold, 2017singh_feature_STIP_graphs_Anomaly_SVM, 2017_kumar_vehicle_object, 2017_Chong_Autoencoder, farooq2017unsupervised, 2017_hinami_Feature_Auto_Model_FasterRCNN, 2017_xu_Feature_SDAE_AnomalyOCSVM, 2016_lan_real_Feature_HOG_Heuristic_hybrid_anomalyusing_relativemotion, 2016_abdallah_feature_3_Model_Hybrid, 2016_zhang_feature_STG_OFM_Model_HOG_SVDD, 2016_fang_Feature_MHOF_SI_PCANet, 2016_patil_global_HOFO_OCSVM,
zhang2016abnormal_Sparsecode, tan2016fast_Sparsecode, 2016_chen_Supervised_Classifier_Feature_OS_SIFT_EML, 2016_zhou_spatial_supervised_Automaticfeatures_STC, 2016_Hasan_Avenue_dataset, 2016_ghrab_abnormal_trajectoryfeature, 2016_medel_Feature_Auto_Model_LSTM, xu2015learning_multifeature_SVM, 2015_nGan_DPMM_outlier, 2015_kaltsa_swarm_Feature_HOG_HOS_PSO_Anomalyby_SVDD, 2015_Cheng_Regression,
2015_yuan_online_feature_SCD_Anomalyby_EMDandMotiondifference, 2015_nguyen_bayesian_feature_Optflowvector_active_iHMM_PCA_BN,
xiao2015learning_unsupervised, chen2015detecting_Hybrid_technique, 2015_ullah_traffic_hybrid_Thermodynamics, 2015_sabokrou_real_STC_GaussianClassifier, 2015_Kruthiventi_dominant_CRF_Clustering_distance, 2015_Xia_vision, 2014_rasheed_tracking_Opticalflow_neuralnetwork, 2014_blair_event_Bayesean, 2017_Sabokrou_DNN, 2014_wang_detection_OCSVM, 2014_mo_adaptive_Sparse_hybrid, 2014_mo_adaptive_Sparse_hybrid, 2014_akoz_traffic_HMM_SVM_Nearest_neibour, 2014_lee_hierarchical_Heuristic_hybrid, 2014_zhu_sparse_Histogram_of_Optical_Flow_Sparsecoding_reconstructioncost,2014_jeong_two_trajectory_hybridlearningLDA_GMM_probability_threshold_measure,
2014_multi_hu_Trajectory_hrybrid_Motionpatternlearning_anymaybymotionpatternmatching,
2014_Yun_motion_MIF_Heuristic, 2014_li_rapid_lowlevelfeature, 2014_elahi_computer_object_and_motion_features,2014_lim_isurveillance_objectfeature, 2014_nallaivarothayan_mrf_stc,2013_Isaloo_Semisupervised_SVM_Trajectory, 2013_li_abnormal_hybrid_features, 2013_kwon_scene_hybrid_features_and_methods,
2013_Yang_trasmil, 2013_Roshtkhari_line
}\\ 
Soft-real time & \cite{sultani2018real_supervised, 2018_DSingh_Autoencoder,  2018_Kaltsa_multiple_HDP, Li2016_ImageDescriptor_GMM, 2016_zhang_video_feature_opticalflow_histogram_STC_method_heuristic_LSH_anomaly_PSO, 2015_pathak_Feature_HOG_HOF_pLSA_anomaly_projectionmodel_histogrambased, 2015_wang_detection_Opticalflow_HOFO_SVM_PCA, 2015_kaviani_automatic_STCs_Hybrid_LDA_pLSA_FSTM, 2015_ranjith_anomaly_Technique_Clustering, 2014_batapati_video_SVM_TRAJECTORY,
2014_kaviani_incorporating_stc, 2013_CLiu_Sparse,
2013_Chockalingam_AVSS
}\\
Offline & \cite{2018_Chang_video_Heuristic,  2017_lin_tube_trajectorybased, 2016_Maiorano_Roughset}\\
\hline
\end{tabular}
\end{table}
\section{Critical Analysis}
\label{sec:Discussion}
This discussion is purely in the context of visual surveillance. Though most of the papers discussed in this survey address anomaly detection, we have observed four key issues with these methods: (i) Benchmark dataset-based comparisons are used to show the effectiveness against the state-of-the-art~\cite{2018_Wang_OCSVM_MotionDescriptor, 2017_Ravanbakhsh_Feature_Normal_OptFrames_Model_GAN_Anomaly_Reconstruction, 2013_CLiu_Sparse, 2017_xu_Feature_SDAE_AnomalyOCSVM}. Though benchmarks may be relevant for comparisons, they may not contain all real-life situations. For example, though anomaly detection works fine on Avenue~\cite{2016_Hasan_Avenue_dataset} dataset, it gives higher false alarms when applied on a real dataset QMUL~\cite{2012_hospedales_video_QMUL_dataset} using two of the proposed methods~\cite{2013_CLiu_Sparse, 2017_Chong_Autoencoder}. Therefore, we believe, the methods need to be relevant for real-life scenarios and should be applicable to long duration videos. 
(ii) Secondly, due to the aforementioned trend, very limited amount of  research~\cite{sultani2018real_supervised, 2018_DSingh_Autoencoder, 2016_chen_Supervised_Classifier_Feature_OS_SIFT_EML} have been carried out for developing generic techniques applicable to a variety of datasets. (iii) There has been hardly any illumination independent research~\cite{2018_DSingh_Autoencoder, 2014_Yun_motion_MIF_Heuristic} except for accident-type anomaly detection. The problem is not entirely due to the limitations of the learning models. It is equally dependent on the dataset types and lack of illumination independent feature extraction. Possibly with the emergence of DNN-based modeling, we hope to address  these issues in future. An object oriented approach might yield better results than histogram-based approaches as human do not think of pixels and their motion in detecting anomalies, but with mere object motion observations. Researchers can make datasets containing segments of the same scene at varying illumination conditions.  (iv) Some approaches remove the background and focus on foreground features for anomaly detections~\cite{farooq2017unsupervised, 2016_lan_real_Feature_HOG_Heuristic_hybrid_anomalyusing_relativemotion, tan2016fast_Sparsecode}. We think, background information should not  be ignored as anomalies also depend on environmental conditions. For example, chance of accidents on a rainy day is higher than that on a sunny day. Obstructions on roads due to various factors should be taken into consideration while preparing datasets. Very few work has happened on this front~\cite{danilescu2015road_pothole, 2016_lan_real_Feature_HOG_Heuristic_hybrid_anomalyusing_relativemotion}. 
\subsection{Challenges and Possibilities}
\label{sec:Challenges_Possibilities}
Some of the stringent challenges on video-based anomaly detection are:
\begin{itemize}
\item \textbf{Illumination}: Even though a handful of anomaly detection methods have already been proposed, the number methods that can handle  illumination variations, are limited~\cite{Li2016_ImageDescriptor_GMM,2015_kaviani_automatic_STCs_Hybrid_LDA_pLSA_FSTM, 2015_Xia_vision}. This is due to the incapabilities of illuminations agnostic feature extraction from the videos. The criteria or methods used under different illumination conditions can be different for real-life applications.
\item \textbf{Pose and Perspective}: Often camera angles focusing on the surveillance area can have substantial impact on the performance of anomaly detection as the appearance of vehicle may change depending on its distance from the cameras~\cite{2015_B_Tian, 2016_Guezouli_Pose, Nakajima2017}. Though object detection accuracy has increased manifolds using deep neural network based methods, still there are challenges in tracking smaller objects. Humans can detect objects at different poses with ease, while machine learning may face difficulties in detecting and tracking the same object under pose variations. 
\item \textbf{Heterogeneous object handling}: Anomaly detection frameworks are largely based on modeling the scene and its entities~\cite{2014_Biswas_HMM, 2018_wang_Modeling_HMM, 2001_Scholkopf_OCSVM, 2015_Cheng_Regression, 2015_sabokrou_real_STC_GaussianClassifier, 2016_Hu_Deep_Features, 2016_medel_Feature_Auto_Model_LSTM, 2013_Yang_trasmil, sultani2018real_supervised, 2014_jeong_two_trajectory_hybridlearningLDA_GMM_probability_threshold_measure, 2015_kaviani_automatic_STCs_Hybrid_LDA_pLSA_FSTM}. However, modeling heterogeneous objects in a scene or learning the movement of heterogeneous objects in a scene can be difficult at times.
\item \textbf{Sparse vs. Dense}: The methods used for detecting anomalies in sparse and dense conditions are different. Though some of the methods~\cite{2013_CLiu_Sparse, 2017_Chong_Autoencoder} are good at locating anomalies in sparse condition, dense scene-based methods can generate many false negatives.
\item \textbf{Curtailed tracks}: Since many anomaly detections are based on vehicle trajectories~\cite{2018_SA_Ahmed, 2016_Maiorano_Roughset, 2014_Biswas_HMM, 2017_Cosar_Pixel_trajectory_Unsupervised, 2013_Yang_trasmil}, underlying tracking algorithms are supposed to perform accurately. Even though tracking accuracies have increased in the last decade, many of the existing tracking algorithms do not work under different scenarios~\cite{2015_Ojha, 2015_B_Tian}. Tracking under occlusion is also another challenge though humans can easily track them visually.
\item \textbf{Lack of real-life datasets}: There is a need for real-life datasets to see the effectiveness of anomaly detection techniques.
\end{itemize}

There are ample scopes and requirements for anomaly detection research based on the gaps discussed earlier. With the advancements in machine learning techniques and affordable hardware, computer vision-based behavior analysis, anomaly detection and anomaly prediction can leapfrog in the coming years. Deep learning-based hybrid frameworks  can handle diverse traffic scenarios. This can also help to build fully automatic traffic analysis frameworks capable of reporting events of interest to the stakeholders.

\section{Conclusion}
In this paper, we have revisited important computer vision-based survey papers. Then, we explored various anomaly detection techniques that can be applied for road network entities involving vehicles, people, and their interaction with the environment. We treat anomaly detection by taking data as the primary unit detailing the learning techniques, features used in learning, approaches employed for anomaly detection, and  applied scenarios for anomaly detection. We intend to set a few future directions  by looking into the gaps in the current computer vision-based techniques  through discussions on various possibilities.
\label{sec:Conclusion}


%

%


\ifCLASSOPTIONcaptionsoff
  \newpage
\fi



\bibliographystyle{plain}

%
%
%

%

\begin{IEEEbiography}[{\includegraphics[width=1in,height=1.25in,clip,keepaspectratio]{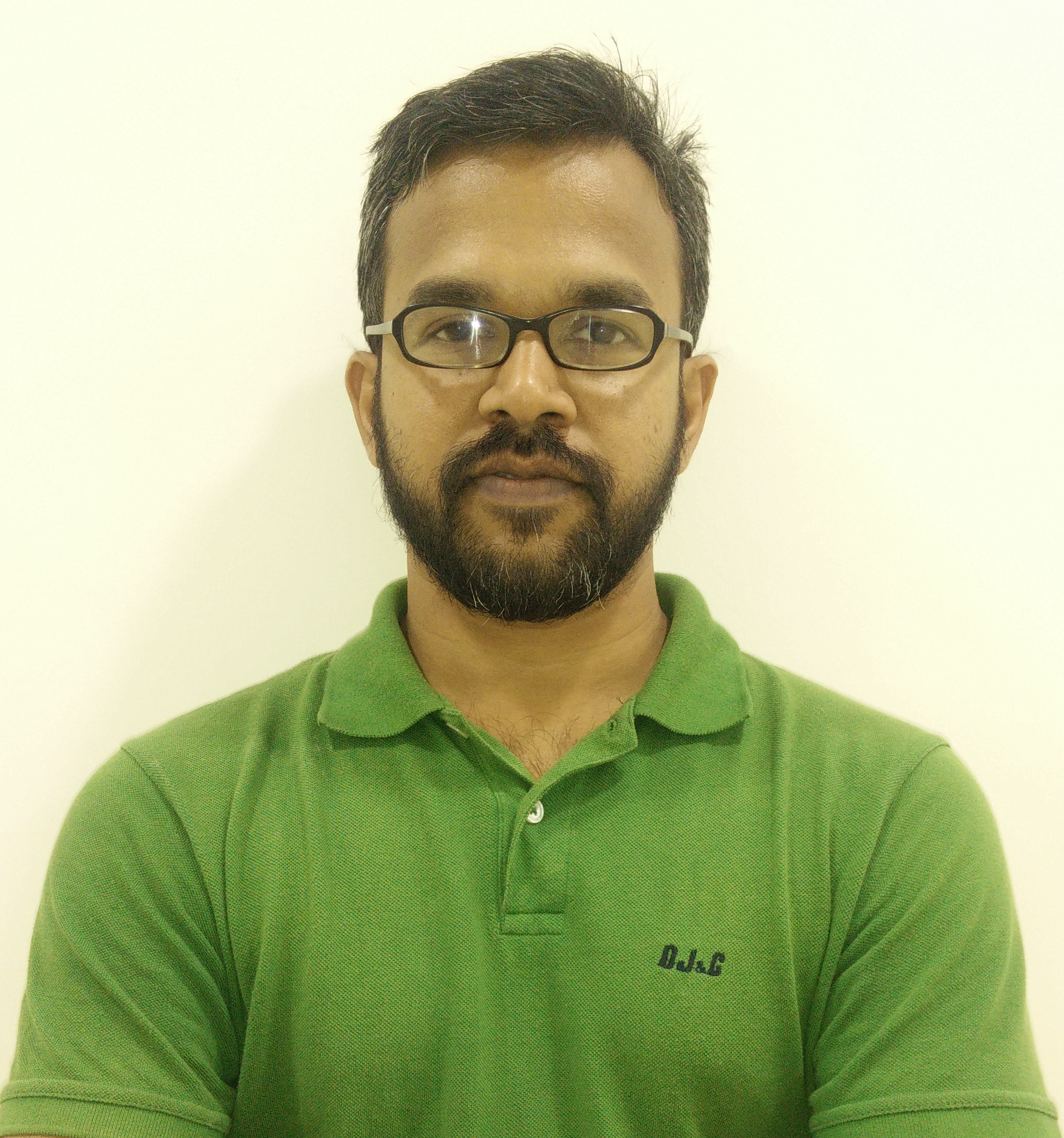}}]{Kelathodi Kumaran Santhosh} is a research scholar in the School of Electrical Sciences, IIT Bhubaneswar, India. He joined a Ph.D. program for resuming his research work that can help humanity. His interests are in the development of vision based applications that can replace human factor. He is a member of IEEE. Prior to joining IIT Bhubaneswar, he worked for Huawei Technologies India Pvt. Ltd. for 10 years (2005-2015) and in Defence Research Development Organization (DRDO) as a Scientist for around 2 years (2003-2004). During his tenure with Huawei, he has worked in many signalling protocols such as Diameter, Radius, SIP etc. in the role of a developer, technical leader, project manager and also served the product lines HSS, CSCF etc. in Huawei China as a support engineer for closer to 1.5 years. In DRDO, he worked in the field of object tracking algorithms based on the data received from radars. More information on Santhosh can be found at https://sites.google.com/site/santhoshkelathodi.
\end{IEEEbiography}
\begin{IEEEbiography}[{\includegraphics[width=1in,height=1.25in,clip,keepaspectratio]{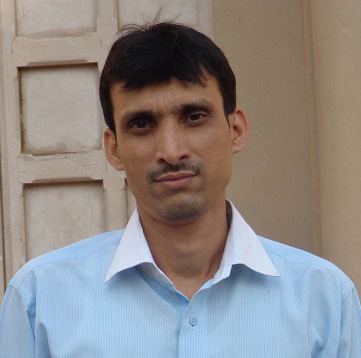}}]{Dr. Debi Prosad Dogra} is an Assistant Professor in the School of Electrical Sciences, IIT Bhubaneswar, India. He received his M.Tech degree from IIT Kanpur in 2003 after completing his B.Tech. (2001) from HIT Haldia, India. After finishing his masters, he joined Haldia Institute of Technology as a faculty members in the Department of Computer Sc. \& Engineering (2003-2006). He has worked with ETRI, South Korea during 2006-2007 as a researcher. Dr. Dogra has published more than 75 international journal and conference papers in the areas of computer vision, image segmentation, and healthcare analysis.  He is a member of IEEE. More information on Dr. Dogra can be found at \url{http://www.iitbbs.ac.in/profile.php/dpdogra}.
\end{IEEEbiography}
\begin{IEEEbiography}[{\includegraphics[width=1in,height=1.25in,clip,keepaspectratio]{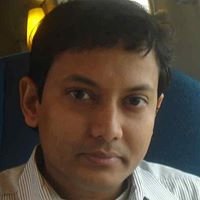}}]{Dr. Partha Pratim Roy} has obtained his M.S. and Ph. D. degrees in the year of 2006 and 2010, respectively at Autonomous University of Barcelona, Spainis. Presently he is an Assistant Professor in the Department of Computer Science and Engineering, IIT Roorkee, India in 2014. Prior to joining, IIT Roorkee, Dr. Roy was with Advanced Technology Group, Samsung Research Institute Noida, India during 2013-2014. Dr. Roy was with Synchromedia Lab, Canada in 2013 and RFAI Lab, France in 2012 as postdoctoral research fellow. His research interests are Pattern Recognition, Multilingual Text Recognition, Biometrics, Computer Vision, Image Segmentation, Machine Learning, and Sequence Classification. He has published more than 100 papers in international journals and conferences. 
\end{IEEEbiography}




\end{document}